\documentclass[sigconf]{acmart}
\AtBeginDocument{%
  }

\setcopyright{acmlicensed}
\copyrightyear{2025}
\acmYear{2025}
\acmDOI{XXXXXXX.XXXXXXX}
\acmConference[Conference acronym 'XX]{Make sure to enter the correct
  conference title from your rights confirmation email}{June 03--05,
  2025}{Woodstock, NY}

\acmISBN{978-1-4503-XXXX-X/2025/10}

\newcommand{\model}{CLEANet}

\usepackage{enumitem}
\usepackage{algorithm}
\usepackage{cleveref}
\crefformat{figure}{Fig.~#2#1#3}
\crefformat{table}{TABLE~#2#1#3}
\crefformat{section}{Section~#2#1#3}
\usepackage{algorithmic}
\usepackage{graphicx}
\usepackage{textcomp}
\usepackage{pifont}
\usepackage[T1]{fontenc}

\usepackage{etoolbox}

\usepackage{multirow}
\usepackage[table,xcdraw]{xcolor}
\usepackage{tcolorbox}
\usepackage{xcolor}
\usepackage{fancybox}
\usepackage{subcaption}
\usepackage{caption}
\usepackage{amsmath} 

\usepackage{amssymb}

\begin{document}
\sloppy

\title{\model: Robust and Efficient Anomaly Detection in Contaminated Multivariate Time Series
}

\author{Songhan Zhang}
\affiliation{%
  \institution{The Chinese University of Hong Kong, Shenzhen}
  \city{Shenzhen}
  \country{China}
}
\email{songhanzhang@link.cuhk.edu.cn}

\author{Yuanhao Lai}
\affiliation{%
  \institution{HUAWEI}
  \city{Shenzhen}
  \country{China}
}
\email{laiyuanhao@huawei.com}

\author{Pengfei Zheng}
\affiliation{%
  \institution{HUAWEI}
  \city{Shenzhen}
  \country{China}
}
\email{zhengpengfei18@huawei.com}

\author{Boxi Yu}
\affiliation{%
  \institution{The Chinese University of Hong Kong, Shenzhen}
  \city{Shenzhen}
  \country{China}
}
\email{boxiyu@link.cuhk.edu.cn}

\author{Xiaoying Tang}
\affiliation{%
  \institution{The Chinese University of Hong Kong, Shenzhen}
  \city{Shenzhen}
  \country{China}
}
\email{tangxiaoying@cuhk.edu.cn}

\author{Qiuai Fu}
\affiliation{%
  \institution{HUAWEI}
  \city{Shenzhen}
  \country{China}
}
\email{fuqiuai@huawei.com}

\author{Pinjia He}
\authornote{Corresponding author.}
\affiliation{%
  \institution{The Chinese University of Hong Kong, Shenzhen}
  \city{Shenzhen}
  \country{China}
}
\email{hepinjia@cuhk.edu.cn}

\renewcommand{\shortauthors}{Songhan Zhang, et al.}

\begin{abstract}
Multivariate time series (MTS) anomaly detection is essential for maintaining the reliability of industrial systems, yet real-world deployment is hindered by two critical challenges: training data contamination (noises and hidden anomalies) and inefficient model inference. Existing unsupervised methods assume clean training data, but contamination distorts learned patterns and degrades detection accuracy. Meanwhile, complex deep models often overfit to contamination and suffer from high latency, limiting practical use.
To address these challenges, we propose \model, a robust and efficient anomaly detection framework in contaminated multivariate time series. \model\ introduces a Contamination-Resilient Training Framework (CRTF) that mitigates the impact of corrupted samples through an adaptive reconstruction weighting strategy combined with clustering-guided contrastive learning, thereby enhancing robustness.
To further avoid overfitting on contaminated data and improve computational efficiency, we design a lightweight conjugate MLP that disentangles temporal and cross-feature dependencies.
Across five public datasets, \model\ achieves up to 73.04\% higher F1 and 81.28\% lower runtime compared with ten state-of-the-art baselines.
Furthermore, integrating CRTF into three advanced models yields an average 5.35\% F1 gain, confirming its strong generalizability.
\end{abstract}




\maketitle

\section{Introduction}
\label{introduction}
Multivariate time series (MTS) anomaly detection has been widely applied in various industry domains, such as the Internet of Things (IoT), Web services, finance \cite{cook2019anomaly}, etc. For example, large-scale online service systems routinely log billions of time series records to monitor server performance, network traffic, and storage usage, furnishing ample information to examine and troubleshoot service anomalies \cite{dragoni2017microservices,balalaie2016microservices}. However, due to the cost and the scarcity of labeled anomaly data in real industries, recent studies focus on developing MTS techniques with unsupervised learning, while reconstruction-based methods that require no labeled anomalies dominate \cite{xu2018unsupervised,su2019robust,audibert2020usad,tuli2022tranad,xu2021anomaly,qi2022mad,chen2023imdiffusion, lai2024nominality,feng2024sensitivehue,wang2024revisiting}.
Reconstruction-based methods learn to encode and decode normal MTS data during training and, at test or detection time, use the reconstruction error as the anomaly score to raise alarms for any arbitrary input. \textbf{Despite such progress, we argue that significant challenges still exist for unsupervised learning-centric MTS anomaly detection techniques.}    

\textbf{Challenge 1: Data contamination undermines the fundamental assumption underlying unsupervised anomaly detection.}
Unsupervised MTS anomaly detection methods presume clean, anomaly-free training data to learn normal patterns. However, in real-world settings, training data often contains unlabeled anomalies and noises due to labeling difficulties, sensor errors, or early-stage faults \cite{tuli2022tranad,xu2024calibrated}. 
Based on our observations, such contamination typically manifests as:

\emph{Salient contamination:} obvious level shifts or spikes with clear deviations from normal distributions.

\emph{Latent contamination:} intricate, irregular deviations that closely resemble normal patterns. 

\begin{figure}[t]
\centerline{\includegraphics[width=0.48\textwidth]{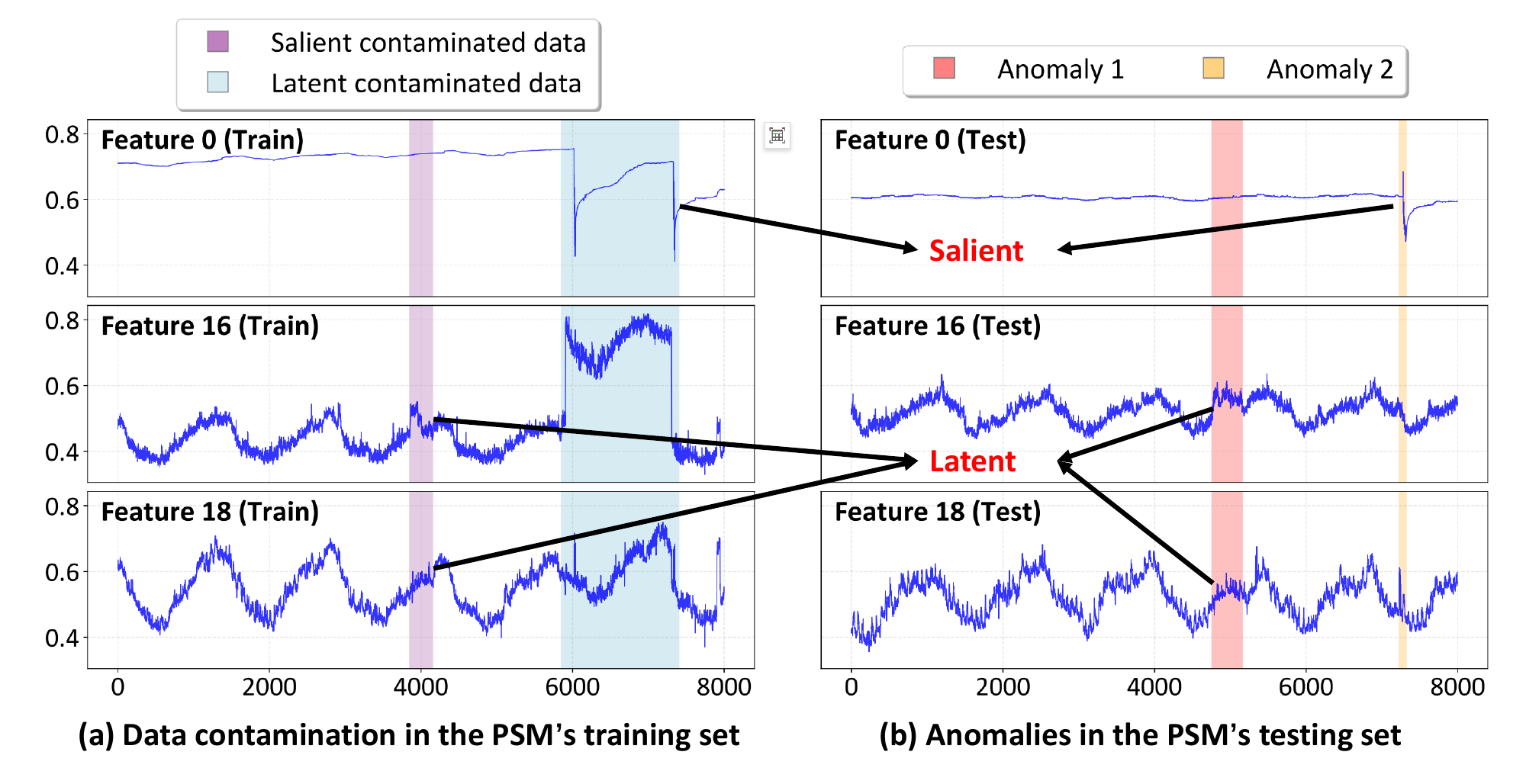}}
\caption{(a) An example of data contamination in the PSM's training set. (b) The evidence of how we identify contaminated data: labeled anomalies in the PSM's testing set.}
\label{Fig0}
\end{figure}

Figure \ref{Fig0} shows a example from the PSM dataset illustrating both types of contamination. We identify them by cross-referencing training samples with testing-set anomaly labels. Salient contamination shows abrupt level shifts, while latent contamination exhibits moderate but misleading deviations without clear spikes. 
Unsupervised learners training on such assumably clean but actually contaminated data can mistakenly learn anomalous patterns as normal, increasing false negatives and hindering generalization. The further these contaminants deviate from the normal distribution, the greater the disruption to the learning process.

Recent work such as TFMAE \cite{fang2024temporal} and TranAD \cite{tuli2022tranad} attempts to mitigate this by injecting artificial noise and using adversial training to enabling robust training against contamination. However, this approach fails to account for existing contamination in the training set because the synthetic noise may not reflect real-world anomalies.
These limitations, along with the underexploration in existing methods motivate us to pursue more robust and principled approaches for contamination-resilient MTS anomaly detection.

\textbf{Challenge 2: Excessive model complexity can hinder anomaly detection performance.} 
Deep learning models with complex architectures, particularly those transformer-based models, have been widely shown to underperform in time series tasks \cite{kim2022towards,kim2024model,lai2024nominality,sun2024unraveling, sarfrazposition,zeng2023transformers, wang2024timemixer, das2304long, fan2024stockmixer, yi2024frequency,yeh2024rpmixer}.
This is because key temporal patterns, like trend and seasonality, can often be effectively captured by simpler linear models.
Moreover, contamination in training data exacerbates this issue, as complex structures are more prone to overfitting such noise, degrading generalization and increasing false negatives.

Recent studies further highlight that the impressive performance of massive state-of-the-art (SOTA) models largely stems from the misuse of an evaluation strategy, Point Adjust (PA) \cite{xu2018unsupervised}, rather than from the models' anomaly detection capabilities \cite{kim2022towards,sun2024unraveling,sarfrazposition}. 
Without PA, simpler methods like Principal Component Analysis (PCA), often outperform those SOTA models \cite{sarfrazposition, sun2024unraveling}. 
These findings caution against the blind pursuit of complexity and highlight the need for models that strike a balance between effectiveness, efficiency, and scalability.

\textbf{Challenge 3: Precise temporal and cross-feature dependencies modeling in moderate-complexity models.}
In MTS, anomalies typically manifest as disruptions in temporal relationships and same-timestamp cross-feature interactions. Prior works in MTS anomaly detection have highlighted the importance of spatiotemporal modeling to effectively capture these dependencies \cite{he2020spatiotemporal, zheng2024graph}.
While complex models, such as those combining Graph Neural Network and Transformer architectures, can naturally model spatiotemporal relationships through intricate computations, this approach is less feasible for moderate-complexity models like MLPs or linear models.
For such lightweight architectures, achieving precise spatiotemporal dependency modeling without relying on excessive computational overhead presents a significant challenge, demanding innovative designs tailored to their simplicity and efficiency.

To address these challenges, we propose \model, a novel \textbf{\underline{C}}ontamination-\textbf{\underline{R}}esilient and efficient multivariate time series \textbf{\underline{A}}nomaly \textbf{\underline{D}}etection approach. \textbf{To ensure robust model training in the presence of data contamination, \model\ proposes a contamination-resilient training framework (CRTF) that systematically trains the model to handle both salient and latent contamination.}
Its core idea is to assign lower reconstruction weights to contaminated data, with the weight decreasing as the degree of contamination increases. To achieve this, we define the degree of contamination based on the consistency of a sample and its local density in the feature space. Building upon this, we propose a novel adaptive weighted reconstruction loss (AWRL) function to mitigate the impact of data contamination in unsupervised training. Furthermore, we design a novel clustering-based contrastive learning method in the latent space to increase the representation distance between normal and contaminated samples, enhancing their separability and improving the robustness of representations.
\textbf{To address the second challenge, \model\ adopts a lightweight MLP structure} to explore a simple yet effective architecture for anomaly detection, and \textbf{further extend it to tackle the third challenge by introducing a conjugate MLP architecture,} consisting of two parallel modules: one for temporal dependencies and the other for cross-feature interactions.
This lightweight structure effectively strikes a balance between functionality and computational efficiency, enhancing robustness to data contamination.

The main contributions of this work can be summarized as follows:

\begin{itemize}[noitemsep, topsep=10pt]

\item We propose \model, a novel contamination-resilient approach for MTS anomaly detection, integrating a robust training framework and a lightweight architecture to ensure effective and efficient anomaly detection.

\item We introduce a contamination-resilient training framework that combines a novel adaptive-weighted reconstruction loss with a clustering-based contrastive learning strategy, effectively mitigating data contamination. 

\item We design a lightweight conjugate MLP architecture with parallel modules for modeling temporal and cross-feature dependencies, reducing inference time by 67\% and model size by 90\% compared to the best baselines.

\item Extensive experiments on five public datasets demonstrate that \model\ outperforms all baselines, achieving an average 73.04\% improvement in F1 score while reducing runtime by 81.28\%.
\end{itemize}

\section{Related Work}
\label{related work}

\textbf{Time Series Anomaly Detection.}
Time series anomaly detection methods can be broadly classified into supervised and unsupervised approaches. \textbf{Supervised methods} such as LSTM-VAE \cite{park2018multimodal}, Spectral Residual \cite{ren2019time}, and CNN \cite{o2015introduction} perform well with labeled data but face limitations due to annotation scarcity.
\textbf{Unsupervised methods} have gained popularity, including statistical techniques (e.g., 3-sigma \cite{blazquez2021review}, ARIMA \cite{box1970distribution}, LOF \cite{breunig2000lof}) and machine learning models (e.g., PCA \cite{mackiewicz1993principal}, KNN \cite{dudani1976distance}, DBSCAN \cite{ester1996density}, iForest \cite{liu2008isolation}, SVM \cite{hearst1998support}), but these struggle with high-dimensional dynamics.

Deep learning-based unsupervised methods, particularly reconstruction-based approaches, dominate time series anomaly detection due to their ability to capture nonlinear patterns automatically. These methods, primarily using Autoencoders (AEs) \cite{sakurada2014anomaly}, encode data into a latent space and reconstruct it, identifying anomalies based on reconstruction errors \cite{xu2018unsupervised,audibert2020usad,su2019robust,li2021multivariate,chen2023imdiffusion}.
USAD \cite{audibert2020usad} enhances robustness with adversarial training, while OmniAnomaly \cite{su2019robust} and InterFusion \cite{li2021multivariate} use RNNs—OmniAnomaly integrates VAEs for temporal modeling, and InterFusion jointly learns inter-metric and temporal embeddings.
Transformer-based methods, such as Anomaly Transformer \cite{xu2021anomaly}, DCdetector \cite{yang2023dcdetector}, TranAD \cite{tuli2022tranad}, and TFMAE \cite{fang2024temporal}, push the field forward by utilizing attention mechanisms. Anomaly Transformer \cite{xu2021anomaly} detects anomalies through association discrepancies, while DCdetector \cite{yang2023dcdetector} enhances local semantics via channel independence patching. TranAD \cite{tuli2022tranad} integrates adversarial training with focus score-based self-conditioning to refine reconstructions, and TFMAE \cite{fang2024temporal} employs dual Transformer autoencoders with temporal and frequency masking to isolate anomalies.
However, massive SOTA models, with their complex architectures, are more prone to overfitting contamination in the training data, which undermines their generalization ability. \model\ addresses this issue by introducing a lightweight and contamination-resilient design.

\textbf{Contrastive Learning.}
Contrastive learning enhances model performance by maximizing similarity between positive pairs while minimizing that of negative pairs \cite{jaiswal2020survey,le2020contrastive}. Originally successful in computer vision (e.g., MoCo, SimCLR, BYOL \cite{he2020momentum,chen2020simple,grill2020bootstrap}), it has been adapted to MTS anomaly detection. Methods like TriAD \cite{sun2024unraveling}, CAE-AD \cite{zhou2022contrastive}, and MGCLAD \cite{qin2023multi} employ data augmentation and SimCLR-based sample pairing. TriAD \cite{sun2024unraveling} injects noise for negative sample construction, while CAE-AD \cite{zhou2022contrastive} incorporates contrastive learning in an LSTM-based autoencoder. However, these approaches often misclassify contaminated data as normal due to unrealistic augmentations, leading to increased false positives.

\model\ overcomes this limitation by proposing a novel clustering-based contrastive learning method, ensuring the effectiveness of positive and negative sample construction and enhancing the robustness of model training.

\section{Methodology}

\begin{figure*}[htbp]
\centerline{\includegraphics[width=1.05\textwidth]{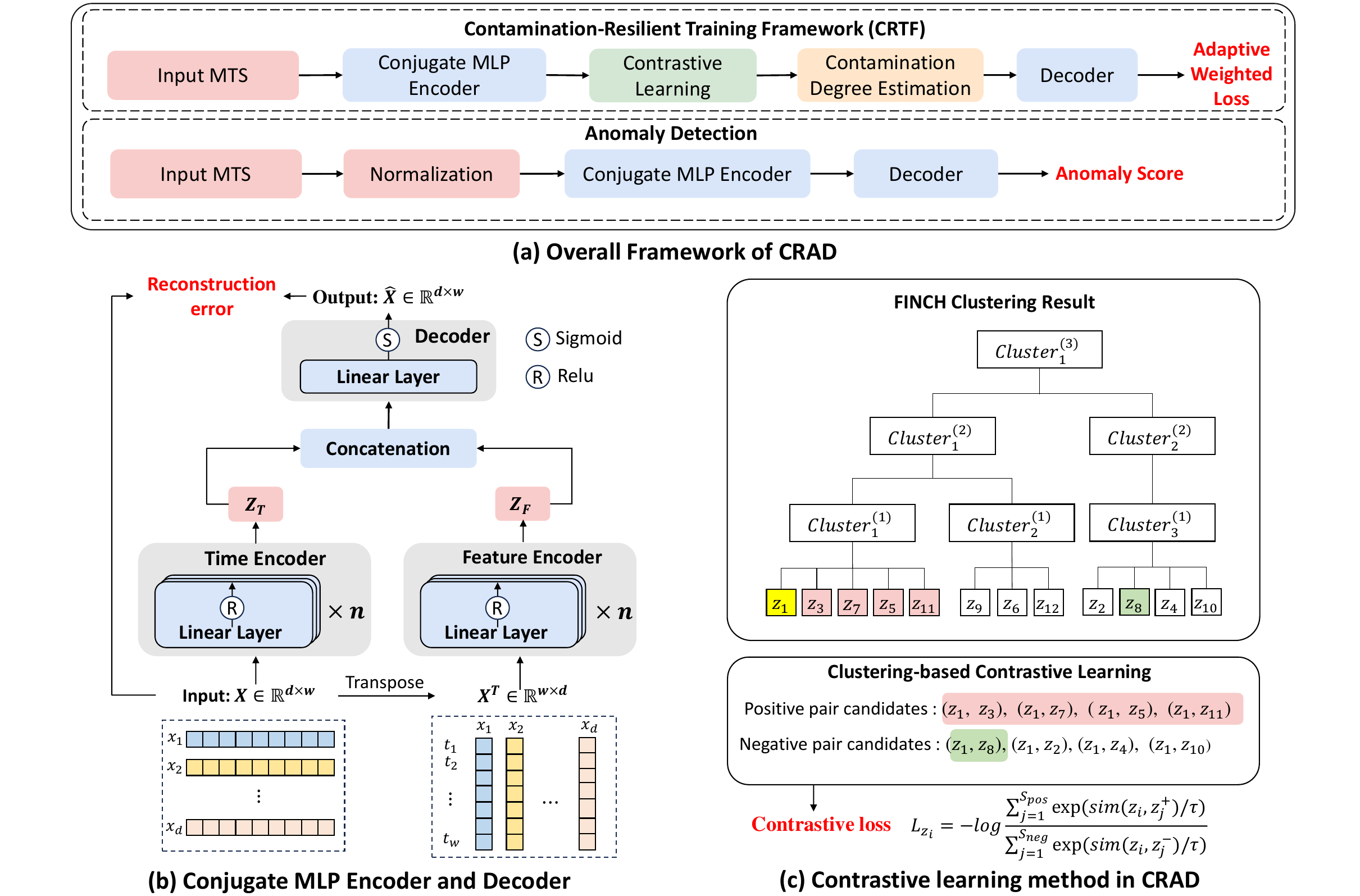}}
\caption{The overview of \model. (a) The overall framework of \model. (b) Conjugate MLP Encoder and Decoder (c) Contrastive learning method in \model.}
\label{Fig2}
\end{figure*}


\subsection{Problem Statement}

\textbf{Multivariate Time Series (MTS)}. A MTS $\textbf{X}\in \mathbb{R}^{d\times T} $ consists of $T$ time points with $d$-dimensional observations at equal intervals, represented by
\begin{equation*}
\centering
\textbf{X}=\left \{ x_{1},x_{2},\dots,x_{T} \right \} 
\end{equation*}
where $x_{i}\in \mathbb{R}^{d}$ denotes a single observation at timestamp $i$ of MTS for $i\in\left [  1,T\right ] $.
It can be further represented by
\begin{equation*}
\centering
x_{i}=(x_{i,1},x_{i,2},\dots, x_{i,d})
\end{equation*}
where $x_{i,j}$ denotes the value of metric $j$ at timestamp $i$.

\textbf{Sliding Window}. A sliding window $X_{t}\in \mathbb{R}^{d\times w}$ is a length-$w$ segment capturing MTS data from time $t-w+1$ to $t$:
\begin{equation*}
\centering
X_{t}=\left \{ x_{t-w+1},x_{t-w+2},\dots,x_{t} \right \}
\end{equation*}

\textbf{Anomaly Detection}. 
Given an unseen window $X_{\text{test}}$, the goal is to predict anomaly labels $\hat{Y} = \left \{ y_1, y_2, \dots, y_w \right \}$ based on an anomaly score. Each $y_{t}\in \left \{ 0,1 \right \} $, where $y_t = 1$ indicates an anomaly if the score exceeds a threshold, and $y_{t} = 0$ indicates a normal point.



\subsection{Conjugate MLP Architecture}
To mitigate the tendency of overly complex models to overfit contaminated data, we propose a conjugate MLP architecture—a lightweight design that remains highly effective.
Unlike existing MLP-based methods \cite{audibert2020usad,xu2018unsupervised} that flatten MTS data into single-dimensional sequences, \model\ directly processes MTS along both temporal and cross-feature dimensions, preserving intra-metric temporal dependencies and inter-metric correlations.
This is crucial as MTS anomaly detection requires capturing cross-feature interactions at each timestamp while analyzing individual metric dynamics, such as trends and seasonality.
Flattening MTS data may overly emphasize asynchronous feature interactions, potentially weakening the model’s ability to detect critical anomalies that manifest as localized deviations or trend disruptions.

We now detail the \model\ architecture. The conjugate MLP encoder consists of two parallel sub-networks: one capturing temporal dependencies from the original data and the other learning cross-feature correlations from its transpose.
Let $L(X) = \sigma(XW+b)$ denote an Single-Layer Perceptron (SLP), where $W$ is the weight matrix and $\sigma$ is the activation function for capturing nonlinearities and accelerating training.
Given an input MTS $X \in \mathbb{R}^{d\times w}$, the time encoder ($TE$) and feature encoder ($FE$) operate as follows:
\begin{equation}
Z_{T}=TE(X),\quad 
Z_{F}=FE(X^{T})
\end{equation}
where $X^{T} \in\mathbb{R}^{w\times d}$ is the transpose of $X$.

An SLP decoder then reconstructs $\hat{X} \in \mathbb{R}^{d\times w}$, ensuring the output retains the same dimensionality as the input. The reconstruction is computed as:
\begin{equation}
\centering
\hat{X}= L_{R}\left (Concat(Z_{T},(Z_{F})^{T})\right ) 
\end{equation}
where $\text{Concat}$ concatenates features along the appropriate dimension. Finally, a sigmoid function normalizes the reconstructed values within $[0,1]$.

\subsection{Contamination-Resilient Training Framework (CRTF)}
\label{contrastive learning}

As discussed in \cref{introduction}, data contamination significantly degrades reconstruction-based MTS anomaly detection methods. To address this, we propose a contamination-resilient training framework (CRTF) that mitigates contamination effects by downweighting contaminated samples and increasing their separation from contaminated ones in the latent space.
CRTF introduces a novel adaptive weighted reconstruction loss (AWRL) that dynamically adjusts sample influence based on contamination level. Additionally, we design a clustering-based contrastive learning strategy, which automatically constructs positive and negative sample pairs to maximize normal-contaminated separation without manual labeling.

\subsubsection{Adaptive Weighted Reconstruction Loss.}
To identify contaminated samples, we define a \textbf{contamination score} $Cont_i$ based on sample' s local consistency and feature-space density. This score quantifies contamination likelihood by jointly measuring (1) similarity to nearby samples and (2) local compactness in the latent space.

Given a dataset $\mathcal{X} = \{ x_1, x_2, \dots, x_n \}$, we compute the representation $z_i$ of sample $x_i$ via the conjugate MLP encoder. The \textbf{consistency score} is:
\begin{equation}
    Cont_i = \frac{1}{|N(i)|} \sum_{j \in N(i)} \text{sim}(z_i, z_j)
\end{equation}
where $N(i)$ denotes the $k$-nearest neighbors and $\text{sim}(\cdot, \cdot)$ is a similarity metric (e.g., cosine similarity). Higher values of $Cont_i$ indicate a greater likelihood that the sample is contaminated, reflecting both semantic deviation and neighborhood sparsity.

\textbf{Adaptive weighting mechanism.}
Existing research has shown that assigning different weights to different samples can enhance model robustness, particularly when dealing with label noise or imbalanced data \cite{ross2017focal,jiang2018mentornet}.
Thus, we define the \textbf{adaptive weight} $w_i$ using a sigmoidal transformation of the contamination score:
\begin{equation}
    w_i = \frac{1}{1 + \exp(\alpha (Cont_i - \tau))}
\end{equation}

where $\tau$ is a contamination threshold and $\alpha$ controls transition sharpness. This ensures lower weights for heavily contaminated samples while preserving information from less contaminated ones.
Finally, the reconstruction loss is redefined as:
\begin{equation}
\centering
L_{AWRL}^{i}=w_i\left \| X_{i}-\hat{X_{i}}  \right \| _{2}^{2}
\end{equation}
enabling robust training under data contamination.

\subsubsection{Clustering-based contrastive learning}
To enhance representation robustness and maximize the distinction between normal and contaminated data, \model\ integrates contrastive learning into the latent space. Contrastive learning, a self-supervised method, encourages similar samples to be closer while pushing dissimilar ones apart \cite{jaiswal2020survey,le2020contrastive}. Existing MTS anomaly detection methods \cite{zhou2022contrastive,qin2023multi,miao2024reconstruction}, inspired by SimCLR \cite{chen2020simple}, typically construct one positive pair via augmentation and treat all others as negatives.
This approach creates misclassification between positive and negative sample pairs, increasing false positives. 
This is because sliding windows in MTS data create overlapping segments, which should form positive pairs but are treated as negatives. Moreover, the periodicity and stationarity of MTS data exacerbate this issue, as recurring patterns increase the likelihood of misclassification.

To address this, inspired by \cite{li2022twin,sharma2020clustering,meng2023mhccl}, we propose a clustering-based contrastive learning method that selects positive and negative pairs based on robust clustering results. This strategy enhances the decision boundary between normal and contaminated data, improving \model's accuracy and robustness.

\textbf{Clustering algorithm}
We employ the FINCH algorithm \cite{sarfraz2019efficient} to cluster samples in the latent space, distinguishing normal from contaminated patterns. 
\model\ begins by calculating the Euclidean distance among samples and then identifies the neighbors of each sample and constructs an adjacency matrix. If $Z_{j}$ is the nearest neighbor of $Z_{i}$, we set $k_{i}^{1}=j$, and the adjacency matrix $A_{ij}$ is defined by
\begin{equation}
\centering
A_{ij}=\begin{cases}
    1, \quad if \;\;  j=k_{i}^{1} \; or \; k_{j}^{1}=i \; or \; k_{i}^{1}=k_{j}^{1}\\
    0, \quad otherwise
\end{cases}
\end{equation}
where $k_{i}^{1}$ denotes the index of the nearest neighbor of sample $i$. \model\ utilizes clusters from the lowest level of this hierarchy to differentiate clean and contaminated patterns.

\textbf{Positive and negative sample pair selection.} For clusters exceeding a predefined threshold $num_{s}$, \model\ constructs positive pairs from the same cluster and negative pairs from the farthest cluster within a minibatch. 
For each anchor sample, \model\ selects positives from its own cluster (if cluster size exceeding a predefined threshold $num_s$) and negatives from the most distant cluster within the same minibatch.
Contaminated clusters are identified by averaging contamination scores across samples. As illustrated in Figure~\ref{Fig2}, traditional methods may mistakenly assign samples from the same cluster as negatives or contaminated ones as positives. In contrast, our method avoids such mismatches by aligning pair construction with semantic consistency in the latent space, thereby improving contrastive training robustness.

\textbf{Contrastive loss.} Assuming that we have $N_{c}$ clusters in the hierarchical clustering result, denoted as $Cluster=\left \{ {C_{1},C_{2},\dots ,C_{N_{c}}}\right \} $. 
For a target cluster $C_{k}$, we denote $(z_{i},z_{i}^{+})$ as a positive pair and $(z_{j},z_{j}^{-})$ as a negative pair. 
The contrastive loss function for each sample is defined by
\begin{equation}
\centering
L_{C_{k}}=-\log\frac{ {\textstyle \sum_{i=1}^{S_{pos}}}exp(sim(z_{i},z_{i}^{+})/\tau^{'} ) }{{\textstyle \sum_{j=1}^{S_{neg}}}exp(sim(z_{j},z_{j}^{-})/\tau^{'} ) }
\end{equation}
where $S_{pos}$ and $S_{neg}$ denote the number of positive and negative samples, $\tau^{'}$ is the temperature parameter, and $sim()$ computes cosine similarity:
\begin{equation}
\centering
 sim(z_{i},z_{j})=\frac{z_{i}\cdot z_{j}}{\left \| z_{i} \right \|\left \|  z_{j}\right \|  } 
\end{equation}

The overall contrastive loss is obtained by averaging across all clusters:
\begin{equation}
\centering
L_{contra}=\frac{1}{K}\sum_{i=1}^{K}L_{C_{k}}  
\end{equation}
where $K$ is the number of clusters used for positive sample construction.

Unlike conventional methods that rely on data augmentation, our clustering-based contrastive learning directly maximizes the representation distance between normal and contaminated samples. This strategy strengthens the decision boundary and improves anomaly detection performance by ensuring robust and stable latent representations.

\subsection{Overvall Framework of CRAD}
During training, \model\ integrates contamination estimation and clustering-based contrastive learning into the latent space of the conjugate MLP encoder, yielding an adaptive weighted loss.
Following a multi-task learning paradigm \cite{zhang2018overview,zhang2021survey}, we jointly optimize reconstruction and contrastive objectives via a weighted sum:
\begin{equation}
\centering
L=L_{AWRL}+\lambda L_{contra}
\end{equation}
where $\lambda$ is a hyperparameter controlling the contribution of contrastive loss.

\textbf{Anomaly Detection.} CRTF is applied only during training. In inference. In inference, given an unseen time window, $X_{\text{test}} =\left \{ x_{\text{test},1}, x_{\text{test},2},\dots,x_{\text{test},w} \right \}$, the anomaly score is computed based on the reconstruction error:
\begin{center}
\begin{equation}
\begin{split}
\hat{W}_{rec} &= L_{R}\left(\text{Concat}(TE(\hat{W}), (FE(\hat{W}^{T}))^{T}\right) \\
&\hspace*{-6pt}\text{Anomaly Score}(\hat{W}) = \left\| \hat{W} - \hat{W}_{rec} \right\| _{2}^{2} \hspace*{\fill}
\end{split}
\end{equation}
\end{center}

If the anomaly score of any time point in $\hat{W}$ exceeds the threshold, this time point is deemed abnormal, represented by:
\begin{equation}
\centering
y_{i}=\begin{cases}
    1, \quad if \;\; w_{i}> threshold\\
    0, \quad otherwise
\end{cases}
\end{equation}
where $y_{i}$ is the predicted label of $w_{i}$. Thus, we obtain the predicted label $\hat{Y}=\left \{ y_{1}, y_{2},\dots,y_{\hat{T}} \right \} $.

\section{Experiment}
\subsection{Experiment Setting}
\textbf{Public Datasets.}
We conduct experiments on five public datasets frequently used in prior studies \cite{audibert2020usad,qi2022mad,paparrizos2022tsb,zhang2023experimental,fang2024temporal}, with statistical details presented in \Cref{table1}. The PSM dataset \cite{abdulaal2021practical} is collected from eBay's application server nodes, containing 26 features and spanning 13 weeks for training and 8 weeks for testing. The SWaT dataset \cite{goh2017dataset,mathur2016swat} simulates an industrial water treatment plant, capturing system operations over 11 days, including anomalies from simulated attacks. The SMD dataset \cite{su2019robust} consists of 5 weeks of server monitoring data from 28 machines, each with 38 metrics. The MSL and SMAP datasets \cite{hundman2018detecting} are collected from NASA spacecraft, recording sensor readings from 27 and 55 entities, respectively.

\begin{table}[t]
\renewcommand{\arraystretch}{1.3}
\setlength\tabcolsep{3.5pt}
\centering
\caption{Statistical information of datasets}
\begin{tabular}{cccccc}
\hline
Dataset & Train  & Test   & Entities & Dimensions & Anomalies(\%) \\ \hline
PSM     & 132481 & 87841  & 1        & 25         & 27.76         \\
SWaT    & 495000 & 449919 & 1        & 51         & 11.98         \\
SMD     & 708405 & 708420 & 28       & 38         & 4.16          \\
MSL     & 58317  & 73729  & 27       & 55         & 10.72         \\
SMAP    & 135183 & 427617 & 55       & 25         & 13.13         \\ \hline
\end{tabular}
\label{table1}
\vspace{-5pt}
\end{table}

For each dataset, we adopt the standardized training and testing sets in the benchmark. 
To implement the \model, we split the original training set into a training set and a validation set in a 4:1 ratio, maintaining the integrity of the testing set.

\textbf{Baseline Approach.}
To demonstrate the superiority of \model, we compare it with 10 baseline methods, including 8 SOTA deep learning models and 2 statistical MTS anomaly detection methods, which have shown advantages in certain scenarios \cite{sarfrazposition}. Since all baselines provide open-source code, we follow their implementations and original parameter settings, selecting optimal thresholds for methods without predefined ones to maximize the F1 score. The deep learning baselines include Omnianomaly \cite{su2019robust}, USAD \cite{audibert2020usad}, Interfusion \cite{li2021multivariate}, Anomaly Transformer \cite{xu2021anomaly}, DCdetector \cite{yang2023dcdetector}, TranAD \cite{tuli2022tranad}, and TFMAE \cite{fang2024temporal} (illustrated in \cref{related work}). Additionally, IMDiffusion \cite{chen2023imdiffusion} explores diffusion models for MTS anomaly detection by leveraging denoising step outputs as anomaly signals. The statistical baselines include PCA, a reconstruction-based dimensionality reduction method, and 1-NN, which detects anomalies based on the nearest neighbor distance \cite{sarfrazposition}.

\textbf{Evaluation Metric.}
We utilize widely-used evaluation metrics to measure the accuracy of anomaly detection:

\begin{equation*}
\centering
P=\frac{TP}{TP+FP}, \quad R=\frac{TP}{TP+FN},\quad F1=2\cdot\frac{ P\cdot R}{P+R} 
\end{equation*}
where TP, FP, and FN denote True Positives, False Positives, and False Negatives, respectively. Precision measures the proportion of correctly predicted anomalies, while Recall quantifies the proportion of detected anomalies among all ground truth anomalies. The F1 score balances both metrics. For the SMD, MSL, and SMAP datasets with multiple entities, we train and test each entity individually and report the average results.

As numerous studies have shown that incorporating the PA strategy often introduces biases into experimental results \cite{kim2022towards,kim2024model,sun2024unraveling}, we exclude the PA strategy from our main experiments (RQ1-RQ4) to ensure a fair evaluation.

\begin{table*}[t]\footnotesize
\renewcommand{\arraystretch}{1.8}
\setlength\tabcolsep{3.5pt}
\centering
\caption{Overall results comparison of \model\ with baseline methods on 5 public datasets. P: Precision, R: Recall, F1: F1 score. The best scores are highlighted in bold, while the second ones are underlined.}
\begin{tabular}{cccccccccccccccc}
\hline
\multirow{2}{*}{\textbf{Method}} & \multicolumn{3}{c}{SWaT}                                     & \multicolumn{3}{c}{SMD}                                      & \multicolumn{3}{c}{PSM}                                      & \multicolumn{3}{c}{MSL}                                      & \multicolumn{3}{c}{SMAP}                                     \\ \cline{2-16} 
                                 & P      & R      & F1                                         & P      & R      & F1                                         & P      & R      & F1                                         & P      & R      & F1                                         & P      & R      & F1                                         \\ \hline
PCA                              & 0.9891 & 0.6081 & \cellcolor[HTML]{F8F8F8}0.7531             & 0.4723 & 0.5381 & \cellcolor[HTML]{F8F8F8}0.5030             & 0.4843 & 0.8048 & \cellcolor[HTML]{F8F8F8}\underline{0.6047} & 0.3827 & 0.6867 & \cellcolor[HTML]{F8F8F8}0.4914             & 0.4357 & 0.4389 & \cellcolor[HTML]{F8F8F8}0.4373             \\
1-NN                             & 0.4618 & 0.3387 & \cellcolor[HTML]{F8F8F8}0.3907             & 0.3524 & 0.6381 & \cellcolor[HTML]{F8F8F8}0.4540             & 0.4334 & 0.8737 & \cellcolor[HTML]{F8F8F8}0.5922             & 0.3782 & 0.4006 & \cellcolor[HTML]{F8F8F8}0.3890             & 0.412  & 0.2582 & \cellcolor[HTML]{F8F8F8}0.3174             \\
OmniAnomaly                      & 0.9783 & 0.6389 & \cellcolor[HTML]{F8F8F8}\underline{0.7730} & 0.4042 & 0.7938 & \cellcolor[HTML]{F8F8F8}\underline{0.5356} & 0.3927 & 0.8696 & \cellcolor[HTML]{F8F8F8}0.5410             & 0.3485 & 0.8349 & \cellcolor[HTML]{F8F8F8}0.4917             & 0.2679 & 0.9084 & \cellcolor[HTML]{F8F8F8}0.4138             \\
USAD                             & 0.9636 & 0.6376 & \cellcolor[HTML]{F8F8F8}0.7674             & 0.3542 & 0.9056 & \cellcolor[HTML]{F8F8F8}0.5092             & 0.4223 & 0.8043 & \cellcolor[HTML]{F8F8F8}0.5538             & 0.4423 & 0.5864 & \cellcolor[HTML]{F8F8F8}\underline{0.5048} & 0.3523 & 0.7078 & \cellcolor[HTML]{F8F8F8}\underline{0.4704} \\
Interfusion                      & 0.8465 & 0.6436 & \cellcolor[HTML]{F8F8F8}0.7312             & 0.3725 & 0.6553 & \cellcolor[HTML]{F8F8F8}0.475              & 0.5384 & 0.5321 & \cellcolor[HTML]{F8F8F8}0.5352             & 0.2386 & 0.9185 & \cellcolor[HTML]{F8F8F8}0.3788             & 0.2137 & 0.7650 & \cellcolor[HTML]{F8F8F8}0.3341             \\
Anomaly Transformer              & 0.1214 & 0.9999 & \cellcolor[HTML]{F8F8F8}0.2166             & 0.0435 & 0.9985 & \cellcolor[HTML]{F8F8F8}0.0834             & 0.3358 & 0.2569 & \cellcolor[HTML]{F8F8F8}0.2910             & 0.1975 & 0.8967 & \cellcolor[HTML]{F8F8F8}0.3237             & 0.1798 & 0.8714 & \cellcolor[HTML]{F8F8F8}0.2981             \\
DCdetector                       & 0.0841 & 0.1065 & \cellcolor[HTML]{F8F8F8}0.094              & 0.1032 & 0.0705 & \cellcolor[HTML]{F8F8F8}0.0838             & 0.0826 & 0.1568 & \cellcolor[HTML]{F8F8F8}0.1082             & 0.1656 & 0.9025 & \cellcolor[HTML]{F8F8F8}0.2799             & 0.1097 & 0.0921 & \cellcolor[HTML]{F8F8F8}0.1001             \\
TranAD                           & 0.9965 & 0.5965 & \cellcolor[HTML]{F8F8F8}0.7463             & 0.6449 & 0.1852 & \cellcolor[HTML]{F8F8F8}0.2878             & 0.4184 & 0.8582 & \cellcolor[HTML]{F8F8F8}0.5625             & 0.2215 & 0.3367 & \cellcolor[HTML]{F8F8F8}0.2672             & 0.1765 & 0.6078 & \cellcolor[HTML]{F8F8F8}0.2736             \\
IMDiffusion                      & 0.6008 & 0.0452 & \cellcolor[HTML]{F8F8F8}0.0841             & 0.7557 & 0.1471 & \cellcolor[HTML]{F8F8F8}0.2332             & 0.1706 & 0.0945 & \cellcolor[HTML]{F8F8F8}0.1216             & 0.3440 & 0.0982 & \cellcolor[HTML]{F8F8F8}0.1528             & 0.0834 & 0.0144 & \cellcolor[HTML]{F8F8F8}0.0245             \\
TFMAE                            & 0.0651 & 0.0023 & \cellcolor[HTML]{F8F8F8}0.0045                                     & 0.1007 & 0.0103 & \cellcolor[HTML]{F8F8F8}0.0187                                     & 0.2959 & 0.0082 & \cellcolor[HTML]{F8F8F8}0.0160                                     & 0.1182 & 0.0100 & \cellcolor[HTML]{F8F8F8}0.0185                                     & 0.1253 & 0.0079 & \cellcolor[HTML]{F8F8F8}0.0149                                     \\ \hline
\textbf{\model}                  & 0.9768 & 0.6792 & \cellcolor[HTML]{F8F8F8}\textbf{0.8013}    & 0.5692 & 0.5137 & \cellcolor[HTML]{F8F8F8}\textbf{0.5401}    & 0.5332 & 0.9113 & \cellcolor[HTML]{F8F8F8}\textbf{0.6728}    & 0.4068  & 0.7430 & \cellcolor[HTML]{F8F8F8}\textbf{0.5257}    & 0.3946 & 0.7415 & \cellcolor[HTML]{F8F8F8}\textbf{0.5150}    \\ \hline
\end{tabular}
\label{table2}
\end{table*}

\subsection{Anomaly Detection Accuracy Result}

The overall performance of the \model\ and baseline methods are detailed in \Cref{table2}.

\begin{figure}[t]
\centerline{\includegraphics[width=0.5\textwidth]{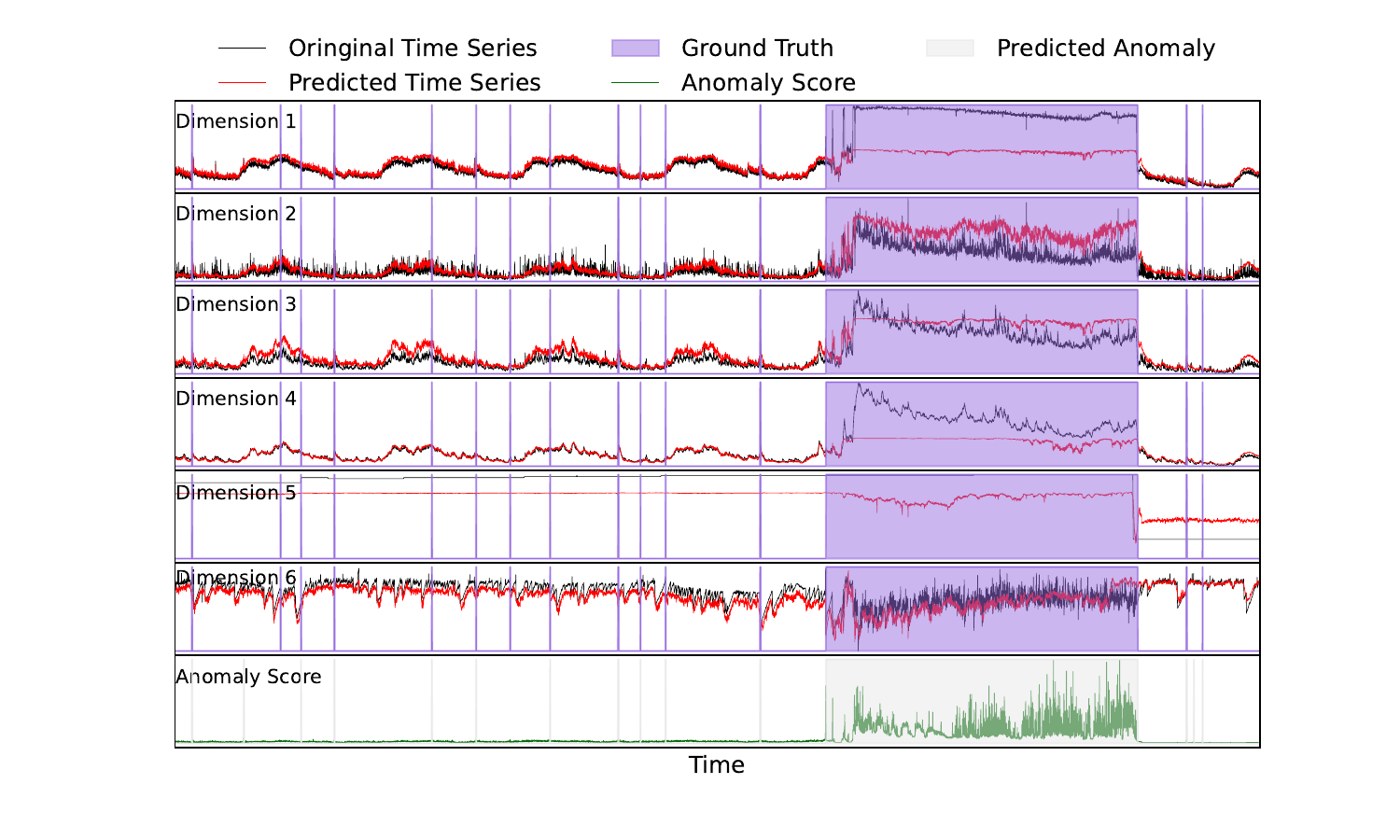}}
\captionsetup{aboveskip=1pt, belowskip=0pt}  
\caption{A sample of anomaly detection results using \model\ on the SMD-Machine-1-6 dataset.}
\label{Fig: sample}
\vspace{-10pt}
\end{figure}

\textbf{\model\ significantly outperforms all baseline methods regarding F1 score across all benchmark datasets,} achieving an average F1 score of 0.6110—an 8.89\% improvement over the second-best method, USAD (0.5611). On the PSM dataset, \model\ surpasses the second-best approach (e.g., PCA) by 11.27\% (absolute increase: 0.0681), which suggests that the CRTF strategy in \model\ enhances the effectiveness and robustness of the model training under data contamination, as the unlabeled anomalies in the PSM's training set are more than those in other datasets.
Furthermore, a sample of anomaly detection results using the \model\ is illustrated in \cref{Fig: sample}, where \model\ demonstrates a strong reconstruction capability and effectively detects salient and most latent anomalies in the ground truth.

\textbf{All methods, except DCdetector and IMDiffusion, achieve their highest F1 scores on SWaT}, likely due to its single-entity structure providing more training data. In contrast, datasets like SMD, MSL, and SMAP consist of multiple entities with varying data quality. For example, \model\ achieves an F1 score of 0.9083 on SMD machine-1-6 but only 0.3228 on machine-1-1, highlighting the necessity of evaluating average performance across entities.

We further analyze why the baseline methods fail to compete with the \model.
The inferior performance of PCA and 1-NN stems from their inability to model nonlinear relationships. Reconstruction-based methods like OmniAnomaly and InterFusion struggle with contaminated training data \cite{su2019robust,li2021multivariate}, making them ineffective for distinguishing anomalies. While adversarial approaches (e.g., USAD \cite{audibert2020usad}, TranAD \cite{tuli2022tranad}) attempt to mitigate this issue, they fail to capture real-world contamination patterns. \model\ overcomes these limitations by introducing CRTF, which enhances the decision boundary between normal and contaminated data.
\model\ overcomes these limitations by introducing CRTF, which enhances the decision boundary between normal and contaminated data.
Additionally, methods like Anomaly Transformer, DCdetector, IMDiffusion, and TFMAE \cite{yang2023dcdetector,xu2021anomaly,chen2023imdiffusion, fang2024temporal} suffer from overfitting to corrupted patterns due to their complex structures or irrelevant components. In contrast, \model's lightweight yet effective architecture enables a more robust understanding of MTS data, leading to superior anomaly detection performance.

\subsection{Anomaly Detection Efficiency Result}

\begin{table}[t]\footnotesize
\renewcommand{\arraystretch}{1.6}
\setlength\tabcolsep{2pt}
\centering
\caption{The efficiency comparison of \model\ with baseline methods on PSM dataset. Unit: Time: s, Parameters: Million, Memory: MiB, MACs: Million. Avg Rank: Average Rank.}
\begin{tabular}{cccccc}
\hline
Method              & Time           & Parameter      & Memory         & MACs           & Avg Rank     \\ \hline
USAD                & 1.339          & 11.917         & 4.197          & 19.843         & 5.5          \\
OmniAnomaly         & 2.198          & 0.904          & 2.099          & 0.625          & 4.8          \\
Interfusion         & 2.137          & 0.217          & 8.125          & 0.599          & 4.0          \\
Anomaly Transformer & 2.998          & 4.802          & 304.565        & 510.720        & 7.8          \\
DCdetector          & 3.977          & 0.836          & 26.227         & 257.760        & 7.0          \\
TranAD              & 1.260          & 0.057          & 1.0815         & 0.326          & 2.0          \\
IMDiffusion         & 92.478         & 0.447          & 42829.514      & 4589.100       & 7.8          \\
TFMAE               & 2.275          & 0.693          & 8.125          & 71.858         & 5.0          \\ \hline
\textbf{\model}     & \textbf{0.428} & \textbf{0.006} & \textbf{1.060} & \textbf{0.325} & \textbf{1.0} \\ \hline
\end{tabular}
\label{table3}
\end{table}

We evaluate the efficiency and deployment performance of \model\ and baseline methods on the PSM dataset. Although tested on a single dataset, the results reflect the model's generalizability, as computation speed and storage primarily depend on model size under a fixed data volume. Following prior works \cite{zeng2023transformers,tuli2022tranad,chen2024lara}, we assess four key efficiency metrics: inference time, memory usage, parameter count, and MACs (Multiple Aggregate Operations), providing a comprehensive view of computational overhead and storage footprint.

\textbf{\model\ achieves superior efficiency across all metrics, setting a new benchmark in performance.} With an inference time of just 0.428 seconds, \model\ is 66\% faster than the second-best method, TranAD (1.260 seconds), making it ideal for real-time anomaly detection. Moreover, \model\ requires only 0.006 million parameters—one-tenth of TranAD’s 0.057 million—highlighting its suitability for resource-constrained environments. In contrast, complex methods like Anomaly Transformer, DCdetector, and IMDiffusion suffer from severe inefficiencies, with IMDiffusion exceeding \model\ in time and memory usage by over 100× and 10,000×, respectively. By combining a lightweight design with high detection accuracy, \model\ strikes an optimal balance between accuracy and efficiency, making it a compelling choice for real-world anomaly detection.

\subsection{Ablation Study}
\label{RQ3}
\begin{table}[t]\footnotesize
\renewcommand{\arraystretch}{1.6}
\setlength\tabcolsep{3.5pt}
\centering
\caption{\model's ablation study across five datasets using F1 score. TE: Time Encoder, FE: Feature Encoder, AWRL: Adaptive Weighted Reconstruction Loss, CL: Contrastive Learning.}
\begin{tabular}{ccccccccc}
\hline
\multicolumn{4}{c}{Variants}               & \multicolumn{5}{c}{Datasets}                                                            \\ \hline
TE      & FE      & AWRL& CL& SWaT            & SMD             & PSM             & MSL             & SMAP            \\ \hline
\multicolumn{4}{c}{\textbf{\model}}        & \textbf{0.8013} & \textbf{0.5401} & \textbf{0.6728} & \textbf{0.5257} & \textbf{0.5150} \\ \hline
\ding{52} & \ding{52} & \ding{52}      & \ding{56} & 0.7893          & 0.5306          & 0.6423          & 0.5173          & 0.5036          \\
\ding{52} & \ding{52} & \ding{56}      & \ding{56} & 0.7704          & 0.5120          & 0.5957          & 0.4972          & 0.4878          \\
\ding{52} & \ding{56} & \ding{56}      & \ding{56} & 0.3567          & 0.3239          & 0.3592          & 0.2307          & 0.2459          \\
\ding{56} & \ding{52} & \ding{56}      & \ding{56} & 0.1246          & 0.0948          & 0.1876          & 0.0894          & 0.0916          \\ \hline
\multicolumn{4}{c}{Flattened MLP}          & 0.7429          & 0.4862          & 0.5765          & 0.4836          & 0.4756          \\ \hline
\multicolumn{4}{c}{\model-SimCLR}          & 0.5354          & 0.3537          & 0.2845          & 0.3503          & 0.3148          \\ \hline
\end{tabular}
\label{table4}
\end{table}

\begin{figure*}[t]
    \centering
    \begin{subfigure}[b]{0.31\textwidth}
        \centering
        \includegraphics[width=\textwidth]{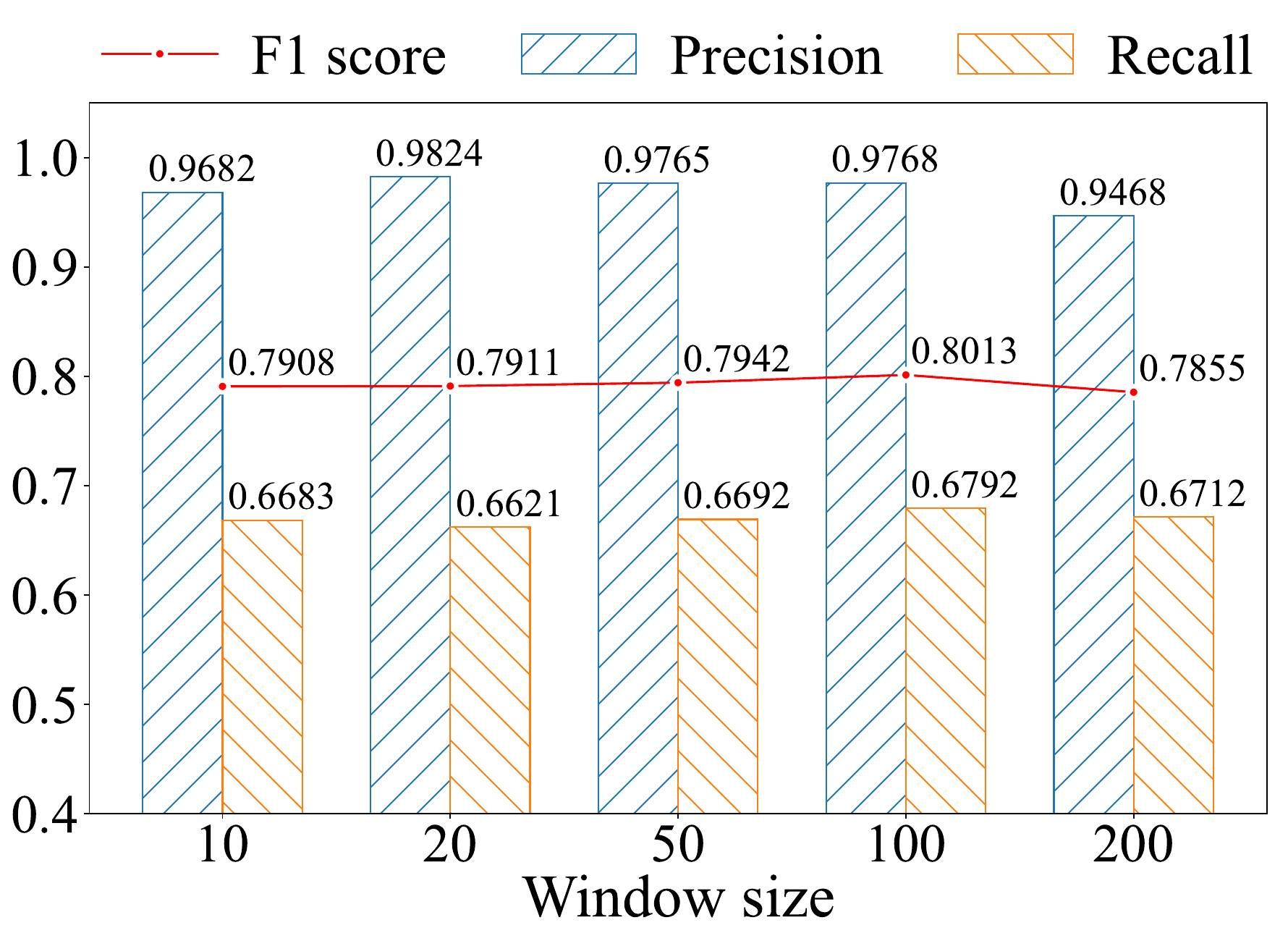}
        \caption{Effect of window size on Accuracy}
        \label{Fig4:subfig1}
    \end{subfigure}
    \hspace{0.01\textwidth} 
    \begin{subfigure}[b]{0.31\textwidth}
        \centering
        \includegraphics[width=\textwidth]{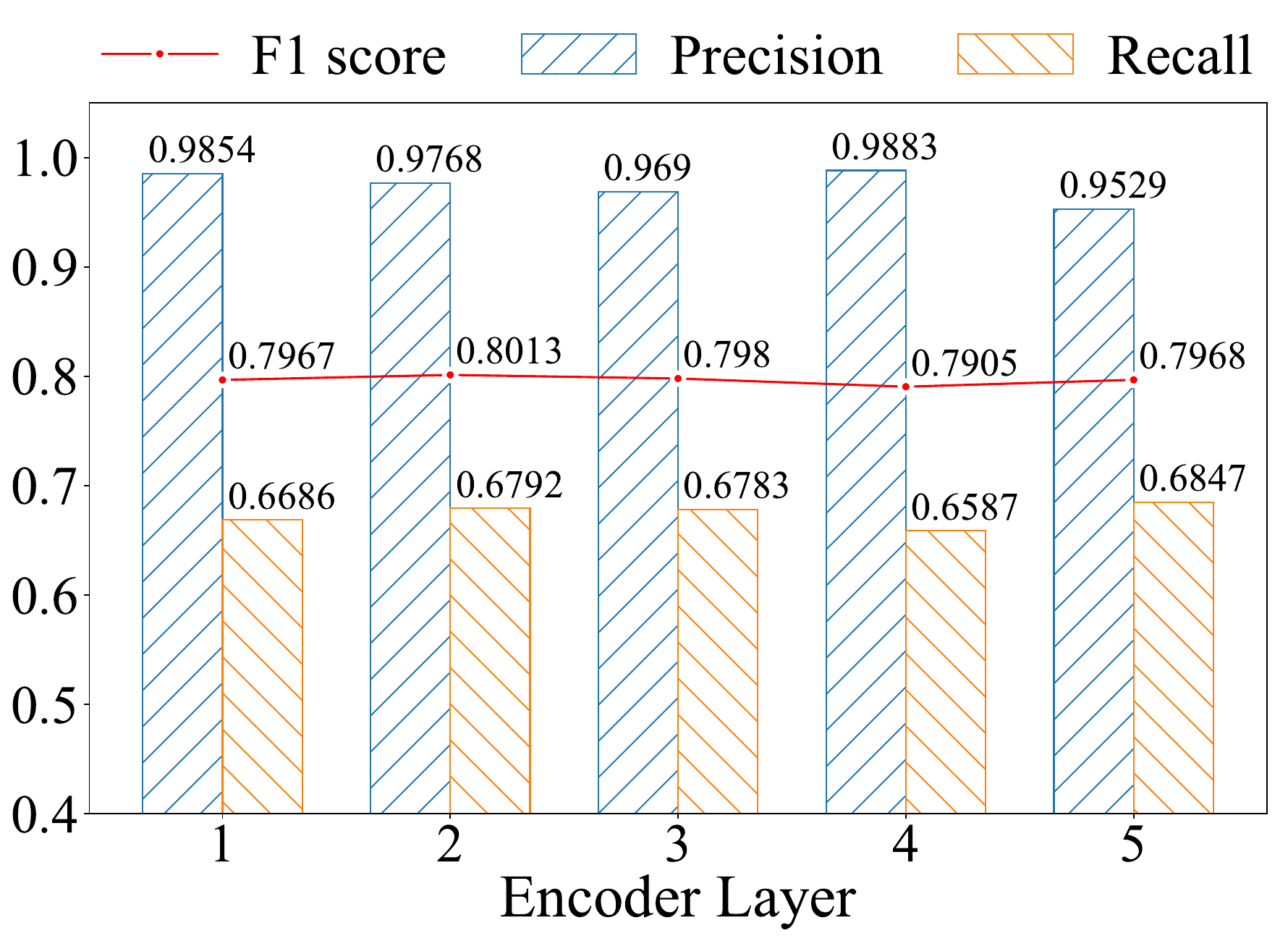}
        \caption{Effect of encoder layers on Accuracy}
        \label{Fig4:subfig2}
    \end{subfigure}
    \hspace{0.01\textwidth}
    \begin{subfigure}[b]{0.31\textwidth}
        \centering
        \includegraphics[width=\textwidth]{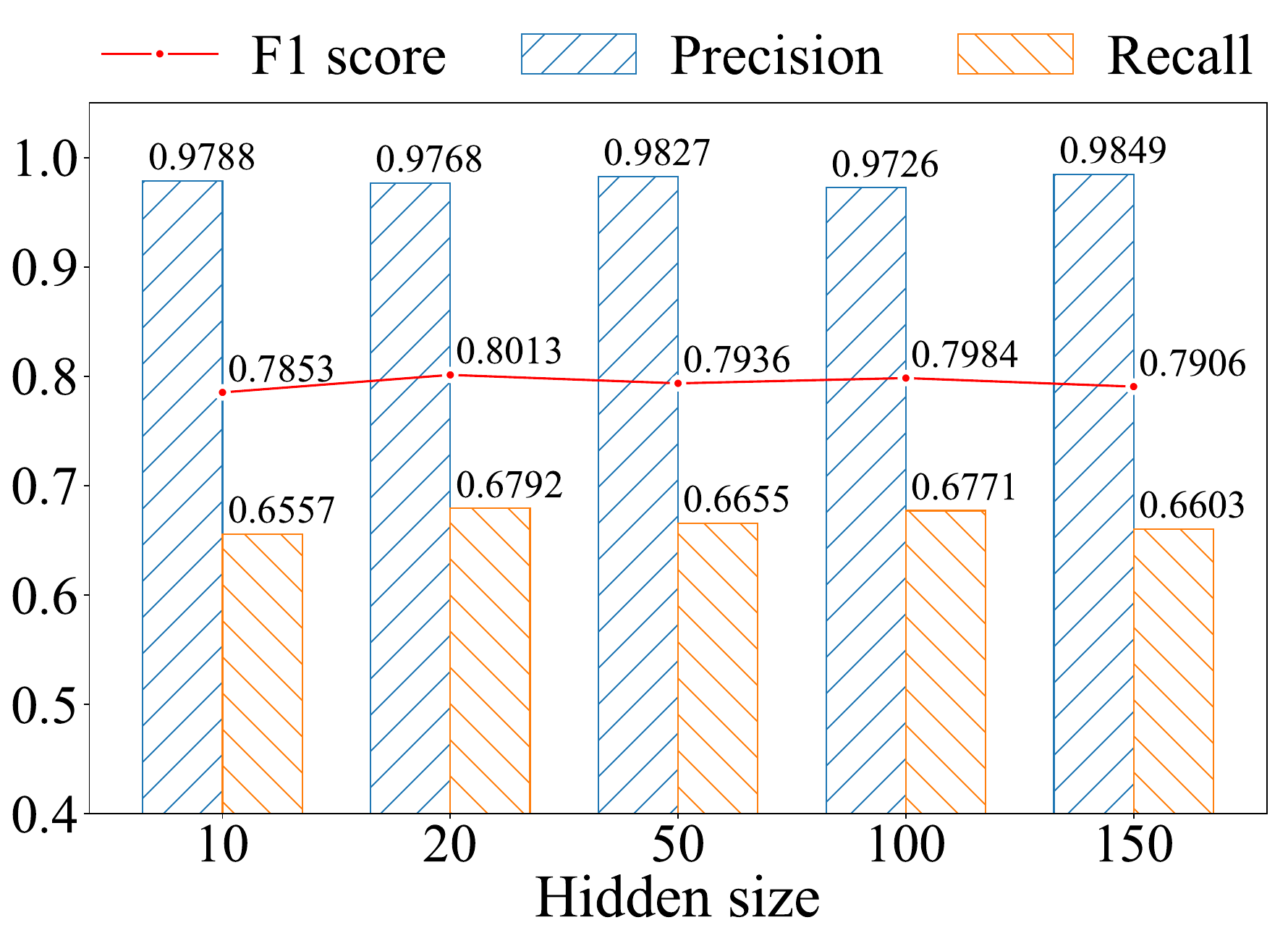}
        \caption{Effect of hidden size on Accuracy}
        \label{Fig4:subfig3}
    \end{subfigure}
    \caption{Parameter sensitivity of window size, number of encoder layers, and hidden size. This analysis is conducted on the SWaT dataset.}
    \label{Fig4}
\end{figure*}

To evaluate the contributions of key components in \model, we conduct an ablation study by systematically omitting each component and assessing its impact on F1 score across five benchmark datasets. First, we investigate the role of the CRTF strategy by removing contrastive learning and replacing adaptive weighted reconstruction loss (AWRL) with a standard reconstruction loss. Second, we examine the effectiveness of the conjugate MLP by designing three variants: a flattened MLP, one without the time encoder, and another without the feature encoder. Finally, we compare our hierarchical clustering-based contrastive learning approach with SimCLR \cite{chen2020simple}, constructing a variant \model-SimCLR to assess its effectiveness in MTS anomaly detection.

\textbf{CRTF significantly enhances \model’s performance, with its removal leading to a 6.25\% average F1 score drop.}
As shown in \Cref{table4}, eliminating contrastive learning reduces the F1 score by an average of 0.0143, while replacing AWRL causes a more substantial decline of 0.0383, with the PSM dataset experiencing the most severe drop (0.0771). These results highlight the critical role of CRTF in mitigating data contamination.
\textbf{Moreover, our proposed contrastive learning method substantially outperforms SimCLR in MTS anomaly detection.}
Replacing it with SimCLR (\model-SimCLR) leads to an F1 score drop of 0.1659–0.3883 across datasets, whereas the model without any contrastive learning still achieves an average F1 score of 0.5726—55.72\% higher than \model-SimCLR. This underscores the limitations of existing MTS contrastive learning approaches in handling complex dependencies.

\textbf{Conjugate MLP significantly improves MTS modeling, boosting F1 score by 10.49\% over the Flattened MLP baseline.}
By jointly modeling temporal and feature dimensions, the conjugate MLP effectively captures multidimensional dependencies in MTS data. Removing either encoder severely degrades performance, with the most dramatic drop (to 0.1076) observed when the time dimension is omitted. This result underscores the necessity of conjugate modeling, as temporal-feature interactions are crucial for capturing intricate patterns in MTS data.

\subsection{Parameter Sensitivity}
In this section, we analyze the impact of key parameters on \model's accuracy using the SWaT dataset, given its extensive data and prevalent anomalies.

\textbf{\model\ demonstrates strong robustness to hyperparameter variations, minimizing the need for extensive tuning.}
As shown in \Cref{Fig4}, its F1 score remains stable across different parameter configurations. The best performance (0.8013 F1-score) is achieved at a window size of 100, yet even with deviations, the model effectively captures temporal patterns. Similarly, varying hidden sizes and encoder layers results in only marginal performance changes, highlighting \model’s adaptability across diverse settings.

\subsection{Generalization Experiment}

\begin{table}[t]
\renewcommand{\arraystretch}{1.3}
\setlength\tabcolsep{1mm}
\centering
\fontsize{9pt}{12pt}\selectfont
\caption{Generalization Experiment of CRTF in \model. w/o CRTF: without CRTF, w CRTF: with CRTF.}
\begin{tabular}{ccccc}
\hline
\multirow{2}{*}{Method} & \multicolumn{2}{c}{SWaT} & \multicolumn{2}{c}{PSM} \\ \cline{2-5} 
                        & w/o CRTF     & w CRTF    & w/o CRTF    & w CRTF    \\ \hline
USAD                    & 0.7674       & 0.7874    & 0.5538      & 0.6180    \\ \hline
OmniAnomaly             & 0.7730       & 0.7826    & 0.5410      & 0.5603    \\ \hline
TranAD                  & 0.7463       & 0.7705    & 0.5625      & 0.6102    \\ \hline
\end{tabular}
\label{append}
\end{table}

To show generalizability of our proposed contamination-resilient training framework (CRTF), we integrate the CRTF into three existing SOTA methods: USAD (MLP-based), OmniAnomaly (RNN-based), and TranAD (Transformer-based).

Table~\ref{append} presents the performance comparison before and after applying the CRTF strategy to existing methods. We observe consistent improvements across all models, with an average F1 score increase of 5.35\%. Notably, USAD and TranAD show significant gains on the PSM dataset, demonstrating the generalizability and effectiveness of our proposed approach in enhancing contamination robustness across diverse architectures.

\section{Conclusion}
This paper presents \model, a contamination-resilient and efficient framework for multivariate time series (MTS) anomaly detection. By introducing a contamination-resilient training framework with an adaptive weighted reconstruction loss and a clustering-based contrastive learning method, \model\ effectively mitigates both salient and latent data contamination.
Additionally, its lightweight conjugate MLP architecture, with parallel modules for temporal and cross-feature modeling, enhances both accuracy and efficiency while reducing overfitting. Extensive experiments on five public datasets show that \model\ outperforms ten baselines, achieving a 73.04\% average improvement in F1 score with significantly lower computational and storage costs.
\model's effectiveness, robustness, and simplicity highlight its potential for real-time anomaly detection across diverse industries.




\end{document}